\documentclass[runningheads]{llncs}
\usepackage[T1]{fontenc}
\usepackage{graphicx}
\usepackage[misc]{ifsym}
\usepackage{amsmath}
\usepackage{amsfonts}

\begin{document}
\title{MedM-VL: What Makes a Good Medical LVLM?}
%
%
\author{Yiming Shi\inst{1}
\and Shaoshuai Yang\inst{1}
\and Xun Zhu\inst{1}
\and Haoyu Wang\inst{1}
\and Xiangling Fu\inst{4}
\and Miao Li\inst{1}$^{(\textrm{\Letter})}$ 
\and Ji Wu\inst{1, 2, 3}$^{(\textrm{\Letter})}$ }
\authorrunning{Y. Shi et al.}
%
\institute{Department of Electronic Engineering, Tsinghua University, Beijing 100084, China\\
\email{\{miao-li, wuji\_ee\}@tsinghua.edu.cn}
\and College of AI, Tsinghua University, Beijing 100084, China\\
\and Beijing National Research Center for Information Science and Technology, Beijing 100084, China\\
\and School of Computer Science, Beijing University of Posts and Telecommunications, Beijing 100876, China}
\maketitle              
\begin{abstract}
Medical image analysis is essential in modern healthcare. Deep learning has redirected research focus toward complex medical multimodal tasks, including report generation and visual question answering. Traditional task-specific models often fall short in handling these challenges. Large vision-language models (LVLMs) offer new solutions for solving such tasks. In this study, we build on the popular LLaVA framework to systematically explore \textbf{model architectures} and \textbf{training strategies} for both 2D and 3D medical LVLMs. We present extensive empirical findings and practical guidance. To support reproducibility and future research, we release a modular codebase, \textbf{MedM-VL}, and two pre-trained models: \textbf{MedM-VL-2D} for 2D medical image analysis and \textbf{MedM-VL-CT-Chest} for 3D CT-based applications. The code is available at: \url{https://github.com/MSIIP/MedM-VL}

\end{abstract}

\section{Introduction}

Medical image analysis plays a critical role in modern healthcare. As clinical needs become increasingly complex, research has shifted toward multimodal tasks~\cite{multimodaltask}, including medical visual question answering (VQA)~\cite{VQA,mm1}, report generation~\cite{mm2,report}, and diagnostic reasoning~\cite{mm3}. These tasks require joint understanding of 2D images (e.g., X-rays), 3D volumes (e.g., CT and MRI), and associated textual information. Early methods rely on task-specific designs or simple fusion, but struggle with complex multimodal integration, cross-modal reasoning, and scalability~\cite{limit1,limit2}. These limitations hinder practical deployment in real-world workflows.

To address these limitations, large vision-language models (LVLMs) have emerged as a promising direction. By leveraging advances in both computer vision and natural language processing, LVLMs aim to unify multimodal understanding within a single framework. These models are particularly well-suited to handle the diversity and complexity of medical data, enabling integrated analysis across 2D and 3D modalities alongside rich textual context~\cite{nature2022multimodal}. Their potential to support open-ended, instruction-driven medical tasks makes them a promising candidate for next-generation medical AI systems.

With the success of ChatGPT-4~\cite{gpt4}, large language models (LLMs)~\cite{llama,qwen} have fundamentally reshaped the landscape of deep learning, enabling powerful capabilities in natural language understanding and generation. Building on this progress, LVLMs~\cite{Qwen-VL,internvl} have emerged as a unified framework for handling complex multimodal tasks that were previously beyond the reach of traditional models. Among them, LLaVA~\cite{llava} stands out for its architectural simplicity and effectiveness. It adopts a standard encoder-connector-LLM architecture and demonstrates that even a linear layer can achieve robust alignment between visual and textual modalities.

Meanwhilie, LVLMs in the medical domain have experienced rapid development~\cite{medflamingo,medgemini,med2e3}, aiming to bridge the gap between complex medical image data and clinical language understanding. LLaVA-Med~\cite{llavamed}, extending the standard LLaVA framework, leverages ChatGPT-4 to generate large-scale instruction-following data, significantly improving the model's ability to handle diverse medical prompts and multimodal tasks. RadFM~\cite{radfm} further advances the filed by training on web-scale multimodal data, encompassing both 2D and 3D medical images, to build a powerful and generalizable foundation model. In parallel, models such as M3D-LaMed~\cite{m3d} and CT-CHAT~\cite{ctrate} focus specifically on 3D CT-based multimodal understanding, addressing the challenges of volumetric data representation and task diversity in diagnostic scenarios.

Although both 2D and 3D medical LVLMs generally adopt the LLaVA-style framework, their model architectures and training strategies differ substantially. A systematic and fair comparison among them remains unexplored. In this study, we introduce MedM-VL, a modular and extensible framework built upon TinyLLaVA Factory~\cite{tinyllavafactory}, tailored for both 2D and 3D medical LVLMs. Based on MedM-VL, we conduct a comprehensive study on how different model architectures and training strategies affect medical LVLM performance. We provide extensive engineering findings and release two pre-trained models: MedM-VL-2D and MedM-VL-CT-Chest. Our main contributions are summarized as follows:
\begin{itemize}
    \item We systematically explore \textbf{model architectures} and \textbf{training strategies} for 2D and 3D medical LVLMs, offering extensive empirical findings and practical guidance.
    \item We release an open-source, modular framework, \textbf{MedM-VL}, to facilitate the training, evaluation, and development of medical LVLMs.
    \item We provide two pre-trained model weights: \textbf{MedM-VL-2D} for 2D medical image analysis and \textbf{MedM-VL-CT-Chest} for 3D CT-based applications.
\end{itemize}

\section{Model Architecture}

\begin{figure}
\includegraphics[width=\textwidth]{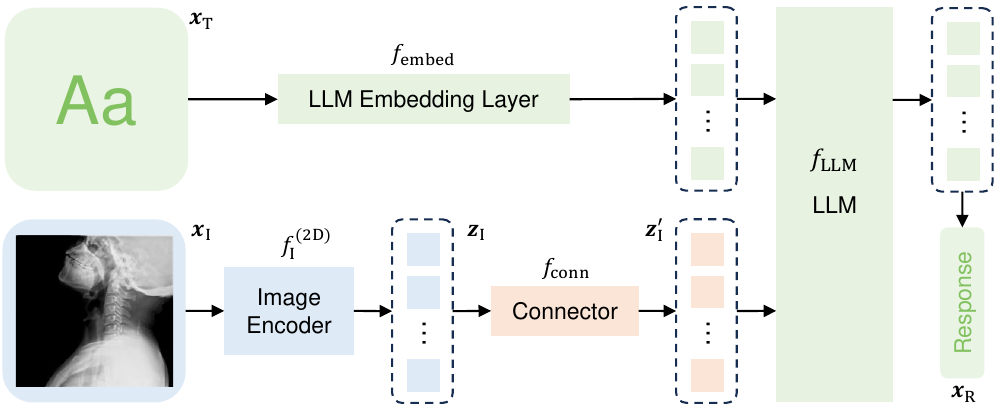}
\caption{Model architecture of MedM-VL-2D. Following the popular LLaVA framework~\cite{llava}, MedM-VL-2D consists of three main components: an image encoder, a connector, and an LLM. It takes a 2D medical image and a textual prompt as input, and generates a textual response, supporting a wide range of medical multimodal tasks.} \label{fig:MedM-VL-2D}
\end{figure}

\subsection{2D Medical LVLMs}
The architecture of 2D medical LVLMs generally follow LLaVA~\cite{llava}, which is one of the most widely adopted architectures for LVLMs. As illustrated in Fig.~\ref{fig:MedM-VL-2D}, it consists of three main components: an image encoder, a connector, and an LLM.

Given a 2D medical image $\mathbf{x}_\text{I}$ and a textual prompt $\mathbf{x}_\text{T}$, the LVLM employs the 2D image encoder $f_{\text{I}}^{\text{(2D)}}$ to extract a sequence of image features $\mathbf{z}_{\text{I}}$:
\begin{equation}
\mathbf{z}_{\text{I}}=f_{\text{I}}^{\text{(2D)}}(\mathbf{x}_{\text{I}})\in \mathbb{R}^{L_{\text{I}}^{\text{(2D)}}\times D_{\text{I}}},
\end{equation}
where $L_{\text{I}}^{\text{(2D)}}$ is the sequence length and $ D_{\text{I}}$ is the embedding dimension.

Next, the connector $f_{\text{conn}}$ projects the image features into the input space of the LLM:
\begin{equation}
\mathbf{z}'_{\text{I}}=f_{\text{conn}}(\mathbf{z}_{\text{I}})\in \mathbb{R}^{L_{\text{I}}^{\text{(2D)}}\times D_{\text{T}}},
\end{equation}
where $D_{\text{T}}$ denotes the embedding dimension of the LLM.

The LLM $f_{\text{LLM}}$ then integrates the image features $\mathbf{z}'_{\text{I}}$ with the textual input $\mathbf{x}_\text{T}$ and generates the textual response $\mathbf{x}_{\text{R}}$:
\begin{equation}
\mathbf{x}_{\text{R}} = f_{\text{LLM}}(\text{concat}(\mathbf{z}'_{\text{I}}, f_{\text{embed}}(\mathbf{x}_{\text{T}}))),
\end{equation}
where the features are concatenated along the sequence (token) dimension, while the embedding dimension remains unchanged.

This LLaVA-style architecture supports multimodal input (image and text) and textual output, enabling a wide range of medical multimodal tasks such as image classification, report generation, VQA, referring expression comprehension (REC), and referring expression generation (REG).

\begin{figure}
\includegraphics[width=\textwidth]{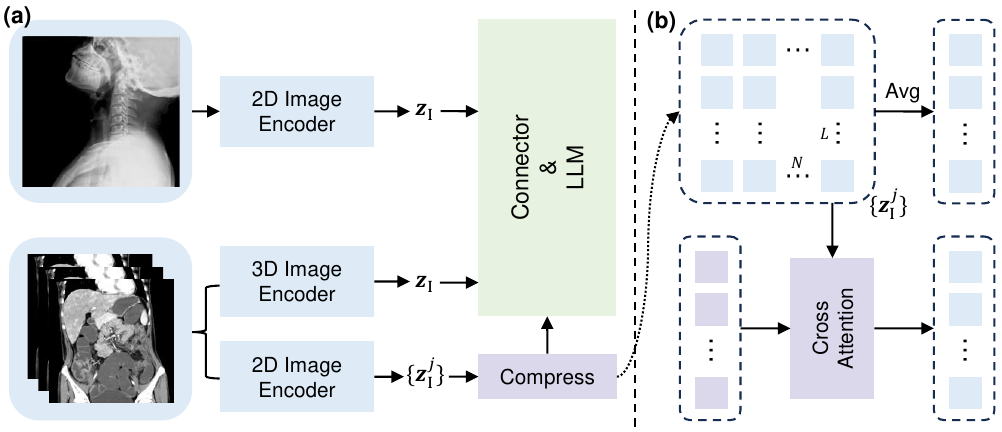}
\caption{(a) Overview of the 3D medical LVLM architecture. Compared to the 2D case, 3D LVLMs adopt either a 3D encoder or apply a 2D image encoder to each slice. Using 2D encoders leads to long feature sequences that require compression. (b) Two compression strategies are explored: a cross-attention module that reduces the sequence to a fixed length, and average pooling across all slice features.}\label{fig:MedM-VL-3D}
\end{figure}

\subsection{3D Medical LVLMs}
The architecture of 3D medical LVLMs largely follows LLaVA~\cite{llava}. However, due to the dimensional differences between 3D and 2D medical images, specialized strategies for feature extraction are required, as illustrated in Fig.~\ref{fig:MedM-VL-3D} (a).

One approach is to directly process the volumetric input $\mathbf{x}_\text{I}^{\text{(3D)}}$ using a 3D image encoder $f_{\text{I}}^{\text{(3D)}}$:
\begin{equation}
\mathbf{z}_{\text{I}}=f_{\text{I}}^{\text{(3D)}}(\mathbf{x}_{\text{I}})\in \mathbb{R}^{L_{\text{I}}^{\text{(3D)}}\times D_{\text{I}}}.
\end{equation}
Alternatively, a 2D image encoder $f_{\text{I}}^{\text{(2D)}}$ can be applied independently to each of the $N$ slices:
\begin{equation}
\mathbf{z}_{\text{I}}^j=f_{\text{I}}^{\text{(2D)}}(\mathbf{x}_{\text{I}}^j)\in \mathbb{R}^{L_{\text{I}}^{\text{(2D)}}\times D_{\text{I}}}, \quad j=1, 2, \dots, N,
\end{equation}
resulting in a long feature sequence of length $N\times L_{\text{I}}^{\text{(2D)}}$. To reduce the sequence length for efficient computation, we explore two compression strategies,as illustrated in Fig.~\ref{fig:MedM-VL-3D} (b).

In the first strategy, all slice-level features are concatenated into a single sequence and passed through a cross-attention module to compress them into a fixed-length sequence of $L_{\text{I}}^{\text{(Attn)}}$ tokens:
\begin{equation}
\mathbf{z}'_{\text{I}}=f_{\text{conn}}\left(\text{concat}(\mathbf{z}_{\text{I}}^1, \mathbf{z}_{\text{I}}^2, \dots, \mathbf{z}_{\text{I}}^N)\right)\in \mathbb{R}^{L_{\text{I}}^{\text{(Attn)}}\times D_{\text{T}}}.
\end{equation}

In the second strategy, an average pooling operation is applied across all slices to produce a compact representation:
\begin{equation}
\mathbf{z}'_{\text{I}}=f_{\text{conn}}\left(\frac{1}{N}\sum_{j=1}^{N}\mathbf{z}_{\text{I}}^j\right)\in \mathbb{R}^{L_{\text{I}}^{\text{(2D)}}\times D_{\text{T}}}.
\end{equation}

Following feature extraction and compression, the LLM integrates image and textual features and generates a textual response:
\begin{equation}
\mathbf{x}_{\text{R}} = f_{\text{LLM}}(\text{concat}(\mathbf{z}'_{\text{I}}, f_{\text{embed}}(\mathbf{x}_{\text{T}}))).
\end{equation}

\section{Training}

\subsection{Data Preparation}
\label{sec:data}

\subsubsection{2D Datasets for Multi-task Learning} The 2D dataset is constructed by integrating multiple public sources to support a wide range of medical multimodal tasks. It includes image classification samples from MedMNIST v2~\cite{medmnist}, report generation data from MIMIC-CXR~\cite{mimic} and MPx-Single~\cite{radfm}, and VQA samples from Path-VQA~\cite{pathvqa} and Slake-VQA~\cite{slake}. For REC and REG, we use data from SA-Med2D-20M~\cite{sa}. To further enhance instruction diversity, we incorporate 60K instruction-tuning samples from the LLaVA-Med~\cite{llavamed} training set.

\subsubsection{3D Dataset Based on Chest CT} Due to the limited availability of 3D medical data, we adopt CT-RATE~\cite{ctrate}, a large-scale 3D medical multimodal dataset consisting of 50K non-contrast chest CT scans from 21K patients. CT-RATE supports various VQA, including long answer, short answer, multiple choice, and report generation. We follow the official training and validation split provided by CT-RATE to ensure consistency.

\subsection{Training Strategy}
We adopt a standard two-stage training strategy~\cite{llava} for both 2D and 3D medical LVLMs. The training process consists of a pre-training stage followed by the instruction-tuning stage.

\subsubsection{Pre-training Stage} In the pre-training stage, the goal is to align the visual and textual modalities. To achieve this, the image encoder and LLM are kept frozen, and only the connector is trained. The training data consists solely of image-caption pairs. 

For the 2D medical LVLM, we use the LLaVA~\cite{llava} pre-training dataset, which contains 558K image-text pairs from the general domain. For the 3D medical LVLM, we use the report generation subset of CT-RATE~\cite{ctrate}.

\subsubsection{Instruction-tuning Stage} The instruction-tuning stage focuses on learning to follow diverse prompts and perform various tasks, aiming to enhance the task generalization ability of LVLMs. In this stage, all model components, including the image encoder, connector, LLM, are trained jointly. The training data includes a wide range of tasks and instruction formats. 

As detailed in Sec.~\ref{sec:data}, the 2D medical LVLM is trained on a multi-task dataset covering image classification, report generation, VQA, REC and REG. The 3D medical LVLM is trained on the full CT-RATE dataset~\cite{ctrate}, which supports long answer, short answer, multiple choice, and report generation.

\section{Experiment}

\subsection{Overall Performance}

\begin{table}
\caption{Training configurations for 2D and 3D medical LVLMs. Values before and after the slash ( / ) indicate settings for 2D and 3D models, respectively.}\label{tab:config}
\centering
\begin{tabular}{|l|c|c|}
\hline
 & Pre-training stage & Instruction-tuning stage\\
\hline
Sequence length & 2048 & 2048 \\
Epoch & 1 & 3/1\\
Batch size & 64/16 & 16/8 \\
Learning rate & 1e-3 & 2e-5 \\
\hline
\end{tabular}
\end{table}

For language compatibility and computational efficiency, we adopt Qwen2.5-3B-Instruct~\cite{qwen2_5} as the LLM. All images or slices are resized to a uniform resolution of $256\times256$. Training is performed using bf16 precision and ZeRO-3 optimization on two NVIDIA A800 GPUs (80GB). 

\begin{table}
\caption{Performance of 2D medical LVLMs across multiple benchmarks.}\label{tab:2d}
\centering
\begin{tabular}{|l|c|c|c|c|c|c|}
\hline
Method & MedMNIST & MedPix & MIMIC-CXR & PathVQA & SAMed & SLAKE\\
\hline
Med-Flamingo~\cite{medflamingo} & 0.089 & 0.081 & \textbf{0.233} & 0.334 & - & 0.215\\
LLaVA-Med~\cite{llavamed} & 0.668 & \textbf{0.151} & 0.204 & 0.378 & 0.458 & 0.337\\
RadFM~\cite{radfm} & 0.189 & - & 0.068 & 0.248 & - & 0.817\\
MedM-VL-2D & \textbf{0.808} & 0.126 & 0.199 & \textbf{0.634} & \textbf{0.693} & \textbf{0.841}\\
\hline
\end{tabular}
\end{table}

For MedM-VL-2D, we use SigLIP~\cite{siglip} as the image encoder and a two-layer MLP as the connector. As shown in Tab.~\ref{tab:2d}, MedM-VL-2D achieves either the best or highly competitive performance across multiple benchmarks, demonstrating its effectiveness in various 2D medical multimodal tasks.

\begin{table}
\caption{Performance of 3D medical LVLMs on CT-RATE~\cite{ctrate}.}\label{tab:3d}
\centering
\begin{tabular}{|l|c|c|c|c|}
\hline
Method & Long & Short & Choice & RG\\
\hline
CT-CHAT~\cite{ctrate} & 0.482 & 0.274 & 0.838 & 0.395\\
MedM-VL-CT-Chest (3D) & 0.619 & 0.658 & \textbf{0.924} & 0.419\\
MedM-VL-CT-Chest (2D+Avg) & 0.622 & 0.664 & 0.920 & \textbf{0.441}\\
MedM-VL-CT-Chest (2D+Attn) & \textbf{0.623} & \textbf{0.667} & \textbf{0.924} & 0.439\\
\hline
\end{tabular}
\end{table}

We compare different strategies for 3D image feature extraction in MedM-VL-CT-Chest. The 3D image encoder is initialized with the pre-trained M3D-CLIP~\cite{m3d}, while the 2D encoder uses the pre-trained SigLIP~\cite{siglip}. All CT volumes are uniformly resized to $32\times256\times256$.

As shown in Tab.~\ref{tab:3d}, despite being pre-trained on large-scale CT datasets covering multiple anatomical regions, the LVLM using the M3D-CLIP encoder underperforms compared to the one using the general-purpose 2D encoder SigLIP. This highlights the strong generalization ability of high-capacity 2D encoders, even when applied to 3D medical volumetric data.

For LVLMs based on the 2D encoder, we evaluate two token compression methods, attention pooling and average pooling. Both approaches achieve comparable results, with attention pooling showing a slight advantage.

\subsection{Ablation Study}

\begin{figure}
\includegraphics[width=\textwidth]{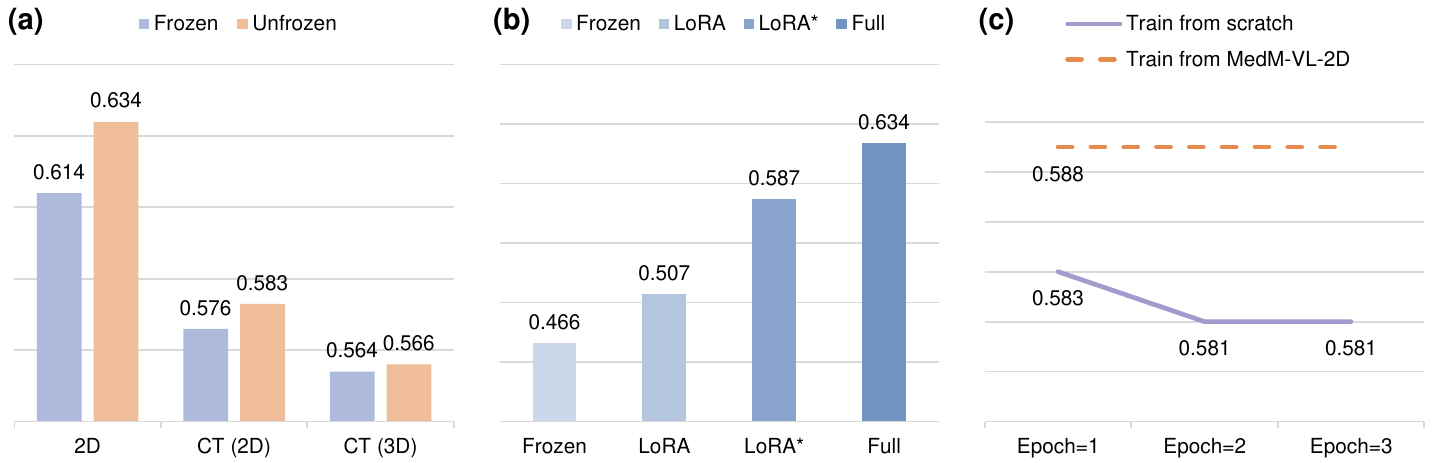}
\caption{(a) Impact of encoder training strategies: frozen vs. unfrozen. (b) Comparison of LLM training strategies: frozen, LoRA, and full-parameter tuning. LoRA* indicates training with a higher learning rate (2e-4). (c) Comparison between training 3D LVLMs from scratch and initializing from pre-trained 2D LVLMs.}\label{fig:trainable}
\end{figure}

\subsubsection{Unfreezing general-purpose encoders improves LVLM performance.} As shown in Fig.~\ref{fig:trainable} (a), unfrozen general-purpose image encoders like SigLIP yield notable gains during instruction tuning, whereas domain-specific encoders such as M3D-CLIP benefit less.

We attribute this difference to the domain gap between the pre-training and fine-tuning datasets. SigLIP, pre-trained on natural images, requires adaptation to medical data, while M3D-CLIP, pre-trained on CT images, is already domain-aligned. This highlights the importance of considering domain alignment when deciding whether to freeze or fine-tune pre-trained encoders in LVLM training.

\subsubsection{Full-parameter tuning of the LLM achieves the best performance.} We compare different LLM training strategies during instruction-tuning, including freezing, LoRA-based tuning, and full-parameter tuning. As shown in Fig.~\ref{fig:trainable} (b), full-parameter tuning yields the best performance, while keeping the LLM frozen results in the worst.

Interestingly, the performance of LoRA is highly sensitive to the learning rate. Increasing the learning rate from 2e-5 to 2e-4 significantly improves results, indicating that appropriate hyperparameter tuning is essential for achieving competitive performance with parameter-efficient methods.

These findings suggest that, while full-parameter tuning remains the most effective, carefully optimized parameter-efficient fine-tuning strategies like LoRA can serve as viable alternatives under resource-constrained settings.

\begin{table}
\caption{Comparison of one-stage and two-stage training strategies. Two-stage training leads to better overall performance.}\label{tab:train_stage}
\centering
\begin{tabular}{|l|c|c|c|c|c|c|}
\hline
Method & MedMNIST & MedPix & MIMIC-CXR & PathVQA & SAMed & SLAKE\\
\hline
One-stage Training & 0.712 & \textbf{0.140} & 0.131 & 0.604 & 0.657 & 0.817\\
Two-stage Training & \textbf{0.808} & 0.126 & \textbf{0.199} & \textbf{0.634} & \textbf{0.693} & \textbf{0.841}\\
\hline
\end{tabular}
\end{table}

\subsubsection{Two-stage training improves LVLM performance.} We compare a one-stage instruction-tuning strategy with a two-stage approach that adds a pre-training stage. As shown in Tab.~\ref{tab:train_stage}, the two-stage strategy achieves superior overall performance. This result highlights the importance of pre-training in aligning visual and textual modalities before exposing the LVLM to diverse task instructions. Without this alignment, instruction-tuning alone may be insufficient for the LVLM to fully exploit the image features, especially in complex medical scenarios.

\subsubsection{3D LVLMs benefit from 2D LVLM initialization.} We compare two strategies for training 3D medical LVLMs: training from scratch and fine-tuning based on a pre-trained 2D medical LVLM. As shown in Fig.~\ref{fig:trainable} (c), initializing the 3D LVLM with weights from a 2D LVLM leads to better performance. Notably, simply extending the training duration of the 3D LVLM does not result in consistent improvement and can even cause performance degradation. This further supports the effectiveness of transferring knowledge from 2D LVLMs, which likely provide a better starting point for multimodal alignment and instruction following.

\section{Conclusion}

In this work, we conduct a comprehensive study of medical LVLMs based on the LLaVA framework. Through systematic comparisons of model architectures and training strategies, we provide practical insights into designing effective medical LVLMs. To support extensibility and reproducibility, we introduce MedM-VL, a modular and flexible codebase that facilitates easy integration and replacement of image encoders, connectors, and LLMs. We hope that our findings and open-source resources will support further research and practical deployment of LVLMs in real-world medical applications.

The field of medical multimodal AI is advancing at a remarkable pace. A diverse array of medical multimodal datasets, encompassing both 2D and 3D images, is continuously emerging. Concurrently, foundational models from the general domain are undergoing rapid progress. Therefore, we believe the pivotal direction for future development lies in how to efficiently leverage existing open-source models and data within resource-constrained environments. Effectively transferring and adapting cutting-edge technologies from the general domain to specific clinical applications will be the key to unlocking the next wave of innovation in the field of medical multimodal intelligence.

\begin{credits}
\subsubsection{\ackname}
This study was funded by Noncommunicable Chronic Diseases-National Science and Technology Major Project (Grant No. 2023ZD0506501), Beijing Natural Science Foundation NO.L251072 and Beijing Natural Science Foundation NO. 4252046.

\subsubsection{\discintname}
The authors have no competing interests to declare that are relevant to the content of this article.
\end{credits}
%
%
%
\bibliographystyle{splncs04}
\bibliography{mybibliography}

\begin{thebibliography}{10}
\providecommand{\url}[1]{\texttt{#1}}
\providecommand{\urlprefix}{URL }
\providecommand{\doi}[1]{https://doi.org/#1}

\bibitem{gpt4}
Achiam, J., Adler, S., Agarwal, S., Ahmad, L., Akkaya, I., Aleman, F.L., Almeida, D., Altenschmidt, J., Altman, S., Anadkat, S., et~al.: Gpt-4 technical report. arXiv preprint arXiv:2303.08774  (2023)

\bibitem{nature2022multimodal}
Acosta, J.N., Falcone, G.J., Rajpurkar, P., Topol, E.J.: Multimodal biomedical ai. Nature medicine  \textbf{28}(9),  1773--1784 (2022)

\bibitem{m3d}
Bai, F., Du, Y., Huang, T., Meng, M.Q.H., Zhao, B.: M3d: Advancing 3d medical image analysis with multi-modal large language models. arXiv preprint arXiv:2404.00578  (2024)

\bibitem{qwen}
Bai, J., Bai, S., Chu, Y., Cui, Z., Dang, K., Deng, X., Fan, Y., Ge, W., Han, Y., Huang, F., et~al.: Qwen technical report. arXiv preprint arXiv:2309.16609  (2023)

\bibitem{Qwen-VL}
Bai, J., Bai, S., Yang, S., Wang, S., Tan, S., Wang, P., Lin, J., Zhou, C., Zhou, J.: Qwen-vl: A versatile vision-language model for understanding, localization, text reading, and beyond. arXiv preprint arXiv:2308.12966  (2023)

\bibitem{VQA}
Ben~Abacha, A., Hasan, S.A., Datla, V.V., Demner-Fushman, D., M{\"u}ller, H.: Vqa-med: Overview of the medical visual question answering task at imageclef 2019. In: Proceedings of CLEF (Conference and Labs of the Evaluation Forum) 2019 Working Notes. 9-12 September 2019 (2019)

\bibitem{limit1}
Bian, Y., Li, J., Ye, C., Jia, X., Yang, Q.: Artificial intelligence in medical imaging: From task-specific models to large-scale foundation models. Chinese Medical Journal  \textbf{138}(06),  651--663 (2025)

\bibitem{internvl}
Chen, Z., Wu, J., Wang, W., Su, W., Chen, G., Xing, S., Zhong, M., Zhang, Q., Zhu, X., Lu, L., et~al.: Internvl: Scaling up vision foundation models and aligning for generic visual-linguistic tasks. In: Proceedings of the IEEE/CVF conference on computer vision and pattern recognition. pp. 24185--24198 (2024)

\bibitem{ctrate}
Hamamci, I.E., Er, S., Almas, F., Simsek, A.G., Esirgun, S.N., Dogan, I., Dasdelen, M.F., Durugol, O.F., Wittmann, B., Amiranashvili, T., Simsar, E., Simsar, M., Erdemir, E.B., Alanbay, A., Sekuboyina, A., Lafci, B., Bluethgen, C., Ozdemir, M.K., Menze, B.: Developing generalist foundation models from a multimodal dataset for 3d computed tomography (2024), \url{https://arxiv.org/abs/2403.17834}

\bibitem{mm1}
Hartsock, I., Rasool, G.: Vision-language models for medical report generation and visual question answering: A review. Frontiers in Artificial Intelligence  \textbf{7},  1430984 (2024)

\bibitem{pathvqa}
He, X., Zhang, Y., Mou, L., Xing, E., Xie, P.: Pathvqa: 30000+ questions for medical visual question answering. arXiv preprint arXiv:2003.10286  (2020)

\bibitem{multimodaltask}
Huang, S.C., Pareek, A., Seyyedi, S., Banerjee, I., Lungren, M.P.: Fusion of medical imaging and electronic health records using deep learning: a systematic review and implementation guidelines. NPJ digital medicine  \textbf{3}(1), ~136 (2020)

\bibitem{tinyllavafactory}
Jia, J., Hu, Y., Weng, X., Shi, Y., Li, M., Zhang, X., Zhou, B., Liu, Z., Luo, J., Huang, L., et~al.: Tinyllava factory: A modularized codebase for small-scale large multimodal models. arXiv preprint arXiv:2405.11788  (2024)

\bibitem{mm2}
Jin, H., Che, H., Lin, Y., Chen, H.: Promptmrg: Diagnosis-driven prompts for medical report generation. In: Proceedings of the AAAI Conference on Artificial Intelligence. vol.~38, pp. 2607--2615 (2024)

\bibitem{mimic}
Johnson, A.E., Pollard, T.J., Berkowitz, S.J., Greenbaum, N.R., Lungren, M.P., Deng, C.y., Mark, R.G., Horng, S.: Mimic-cxr, a de-identified publicly available database of chest radiographs with free-text reports. Scientific data  \textbf{6}(1), ~317 (2019)

\bibitem{llavamed}
Li, C., Wong, C., Zhang, S., Usuyama, N., Liu, H., Yang, J., Naumann, T., Poon, H., Gao, J.: Llava-med: Training a large language-and-vision assistant for biomedicine in one day. Advances in Neural Information Processing Systems  \textbf{36},  28541--28564 (2023)

\bibitem{slake}
Liu, B., Zhan, L.M., Xu, L., Ma, L., Yang, Y., Wu, X.M.: Slake: A semantically-labeled knowledge-enhanced dataset for medical visual question answering. In: 2021 IEEE 18th international symposium on biomedical imaging (ISBI). pp. 1650--1654. IEEE (2021)

\bibitem{report}
Liu, G., Hsu, T.M.H., McDermott, M., Boag, W., Weng, W.H., Szolovits, P., Ghassemi, M.: Clinically accurate chest x-ray report generation. In: Machine Learning for Healthcare Conference. pp. 249--269. PMLR (2019)

\bibitem{llava}
Liu, H., Li, C., Wu, Q., Lee, Y.J.: Visual instruction tuning. Advances in neural information processing systems  \textbf{36},  34892--34916 (2023)

\bibitem{medflamingo}
Moor, M., Huang, Q., Wu, S., Yasunaga, M., Dalmia, Y., Leskovec, J., Zakka, C., Reis, E.P., Rajpurkar, P.: Med-flamingo: a multimodal medical few-shot learner. In: Machine Learning for Health (ML4H). pp. 353--367. PMLR (2023)

\bibitem{mm3}
Savage, T., Nayak, A., Gallo, R., Rangan, E., Chen, J.H.: Diagnostic reasoning prompts reveal the potential for large language model interpretability in medicine. NPJ Digital Medicine  \textbf{7}(1), ~20 (2024)

\bibitem{med2e3}
Shi, Y., Zhu, X., Hu, Y., Guo, C., Li, M., Wu, J.: Med-2e3: A 2d-enhanced 3d medical multimodal large language model. arXiv preprint arXiv:2411.12783  (2024)

\bibitem{limit2}
Sopruchi, A., Anguzu, R., University~II, K.I.: The integration of ai-driven decision support systems in healthcare: Enhancements, challenges, and future directions. IDOSR JOURNAL OF COMPUTER AND APPLIED SCIENCES  \textbf{9},  17--25 (10 2024). \doi{10.59298/JCAS/2024/92.1725}

\bibitem{llama}
Touvron, H., Lavril, T., Izacard, G., Martinet, X., Lachaux, M.A., Lacroix, T., Rozi{\`e}re, B., Goyal, N., Hambro, E., Azhar, F., et~al.: Llama: Open and efficient foundation language models. arXiv preprint arXiv:2302.13971  (2023)

\bibitem{radfm}
Wu, C., Zhang, X., Zhang, Y., Wang, Y., Xie, W.: Towards generalist foundation model for radiology. arXiv preprint arXiv:2308.02463  (2023)

\bibitem{qwen2_5}
Yang, A., Yang, B., Zhang, B., Hui, B., Zheng, B., Yu, B., Li, C., Liu, D., Huang, F., Wei, H., et~al.: Qwen2. 5 technical report. arXiv preprint arXiv:2412.15115  (2024)

\bibitem{medmnist}
Yang, J., Shi, R., Wei, D., Liu, Z., Zhao, L., Ke, B., Pfister, H., Ni, B.: Medmnist v2-a large-scale lightweight benchmark for 2d and 3d biomedical image classification. Scientific Data  \textbf{10}(1), ~41 (2023)

\bibitem{medgemini}
Yang, L., Xu, S., Sellergren, A., Kohlberger, T., Zhou, Y., Ktena, I., Kiraly, A., Ahmed, F., Hormozdiari, F., Jaroensri, T., et~al.: Advancing multimodal medical capabilities of gemini. arXiv preprint arXiv:2405.03162  (2024)

\bibitem{sa}
Ye, J., Cheng, J., Chen, J., Deng, Z., Li, T., Wang, H., Su, Y., Huang, Z., Chen, J., Jiang, L., et~al.: Sa-med2d-20m dataset: Segment anything in 2d medical imaging with 20 million masks. arXiv preprint arXiv:2311.11969  (2023)

\bibitem{siglip}
Zhai, X., Mustafa, B., Kolesnikov, A., Beyer, L.: Sigmoid loss for language image pre-training. In: Proceedings of the IEEE/CVF International Conference on Computer Vision. pp. 11975--11986 (2023)

\end{thebibliography}

\end{document}